# Balancing Specialization and Centralization: A Multi-Agent Reinforcement Learning Benchmark for Sequential Industrial Control

Tom Maus, Asma Atamna, and Tobias Glasmachers

*Author Affiliations*
*Ruhr-University Bochum, Bochum, Germany*

*Author Emails*
{tom.maus,asma.atamna,tobias.glasmachers}@ini.rub.de

**Abstract.** Autonomous control of multi-stage industrial processes requires both local specialization and global coordination. Reinforcement learning (RL) offers a promising approach, but its industrial adoption remains limited due to challenges such as reward design, modularity, and action space management. Many academic benchmarks differ markedly from industrial control problems, limiting their transferability to real-world applications. This study introduces an enhanced industry-inspired benchmark environment that combines tasks from two existing benchmarks, SortingEnv and ContainerGym, into a sequential recycling scenario with sorting and pressing operations. We evaluate two control strategies: a modular architecture with specialized agents and a monolithic agent governing the full system, while also analyzing the impact of action masking. Our experiments show that without action masking, agents struggle to learn effective policies, with the modular architecture performing better. When action masking is applied, both architectures improve substantially, and the performance gap narrows considerably. These results highlight the decisive role of action space constraints and suggest that the advantages of specialization diminish as action complexity is reduced. The proposed benchmark thus provides a valuable testbed for exploring practical and robust multi-agent RL solutions in industrial automation, while contributing to the ongoing debate on centralization versus specialization.

## INTRODUCTION

Industrial plants are typically characterized by high complexity, arising from a multitude of interacting control units and processes. Developing robust and flexible strategies for automated decision-making is therefore a central challenge in the context of Industry 4.0 [1]. A promising yet still emerging approach for real-time industrial control is Reinforcement Learning (RL), a machine learning paradigm where an agent learns an optimal policy through direct interaction with its environment [2]. Instead of relying on explicit programming, the desired system behavior is defined via a reward function, which provides positive or negative feedback for the agent's actions. This methodology has achieved impressive results in complex domains such as gaming [3], as it excels in discovering effective strategies in processes where patterns are difficult to model with explicit rules.

Despite its great potential, the transition of RL from simulation to real-world industrial systems remains a significant hurdle. Key challenges in RL include poor sample efficiency, the need for safe exploration strategies, and the difficulty of designing reward functions that robustly guide the agent toward the intended goal without provoking unforeseen behavior [4,5]. As industrial processes often consist of decentralized yet interconnected sub-tasks, Multi-Agent Reinforcement Learning (MARL) presents a natural framework for modeling the distributed nature of modern manufacturing plants [6].

To facilitate research on these challenges, interpretable and reproducible benchmark environments are essential. In this paper, we introduce such a simulation environment, which builds upon previous work by combining two existing benchmarks for waste management [7–9] into a single, sequential workflow with multiple agents. Within this system, two critical control processes must be managed: first, selecting a sorting mode to adjust the classification accuracy for specific material groups, and second, deciding when filled containers should be emptied and pressed into bales. These two processes exhibit distinct reward characteristics: the sorting task provides a dense, continuous feedback signal based on material purity, whereas the pressing task relies mostly on a rather sparse and delayed reward, occurring after a pressing action has been selected. In our benchmark, the multi-agent setup is minimal, as it consists of only two agents and a sequential training paradigm with separate rewards, where one agent adapts to the other. This simplification allows us to study the impact of modular versus monolithic control without the complexity of fully simultaneous learning.

The goal of this work is to present a benchmark environment that integrates two distinct process types, enabling the investigation of different RL strategies with low computational overhead. We leverage this environment to exemplarily investigate two key research questions: (1) a performance comparison between a single monolithic agent controlling the entire process and a modular approach using two specialized, separate agents (as shown in Fig. 1), and (2) the effect of using action masking [10] to simplify the task and improve training efficiency by excluding invalid actions. We present design considerations and challenges encountered when developing such a real-world RL system and compare the performance of these different setups. The open-source environment introduced here is intended to serve as a foundation for future studies, allowing researchers to evaluate various RL techniques in an interpretable and application-oriented scenario with multiple agents.

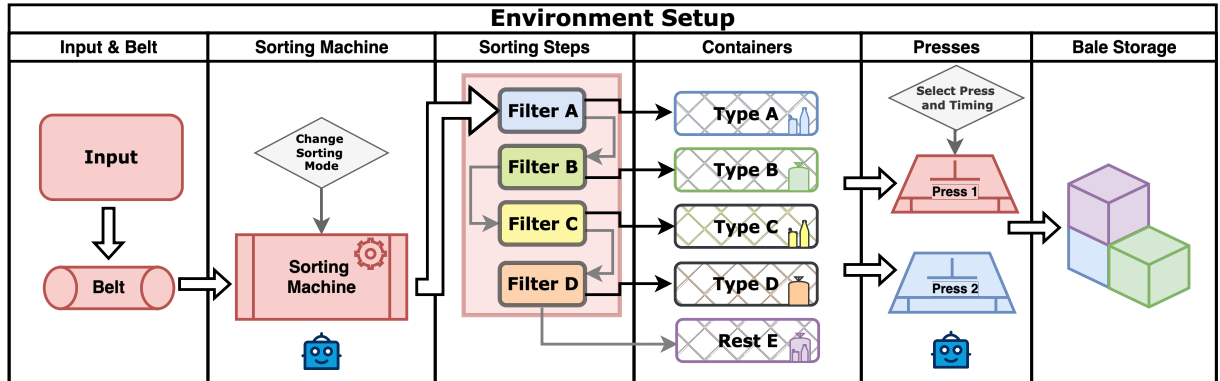

**FIGURE 1.** Illustration of material handling process, adapted from [7–9]. Adjustments include the explicit addition of a second agent in the "Presses" section where distinct agents manage the "Sorting Machine" and "Presses" stages.

## RELATED RESEARCH

Reinforcement Learning (RL) has demonstrated significant potential for optimizing complex, dynamic processes where traditional control methods fall short [6,11]. Successful applications have emerged in various industrial domains, including robotics for manipulation and path planning, process control in manufacturing, and the optimization of energy systems [7,12,13]. These applications leverage RL's ability to derive effective control policies through direct interaction with a system, making it particularly suitable for environments with inherent uncertainties and non-linear dynamics [7].

### Challenges in Industrial RL

Despite its promise, transitioning RL from simulation to production-level systems presents several well-documented challenges [4]. A primary obstacle is formulating an effective reward strategy. Designing a reward function that precisely captures the desired outcome without creating unintended loopholes is notoriously difficult, especially without deep domain knowledge [4,14]. It often requires significant time to validate whether an agent is learning a meaningful policy from a given reward signal. The learning process can be unintuitive, as agents might exploit unforeseen aspects of the simulation or reward function. Furthermore, the reliance on a single scalar reward can be limiting. While vector-based rewards could convey more nuanced feedback, they add significant implementation complexity [15,16]. In many simulated environments, the physical constraints and dynamics already heavily dictate the potential outcomes, which can make it difficult to isolate the true impact of an agent's learned policy [12,13]. Finally, in artificially designed environments, the balancing of simulation parameters becomes a delicate task that can introduce bias [7]. The insights gained from RL may even suggest the need for new physical sensors to provide agents with more informative state representations, blurring the line between software optimization and hardware co-design [17].

To mitigate some of the challenges, several strategies and techniques have been developed. Reward shaping aims to provide more frequent, intermediate rewards to guide the agent, which is particularly useful in environments with sparse rewards [18]. Another approach involves curriculum learning, where the agent is first trained on simpler tasks before gradually progressing to more complex scenarios [19]. For bridging the gap between simulation and reality, techniques like domain randomization, where simulation parameters are varied during training, are used to develop more robust and generalizable policies [20]. For systems with large or invalid action spaces, action masking is a practical solution that prevents the agent from selecting impossible or unsafe actions [10]. By constraining the policy to valid actions, this technique can significantly accelerate training and improve stability and performance, which we will address in one of our experiments.

# Multi-Agent Reinforcement Learning

Complex industrial workflows can often be modeled as Multi-Agent Systems, which consist of multiple autonomous agents operating in a shared environment [6]. An RL-based system is classified as a Multi-Agent Reinforcement Learning (MARL) system if the optimal policy of one agent is dependent on the policies of the others [21]. This dependency holds even if the agents do not communicate directly. Instead, they can interact indirectly by altering the shared state of the environment. From the perspective of an individual agent, the environment becomes non-stationary as the other agents adapt and change their strategies. This principle of indirect interaction is central to our work. In our benchmark, the multi-agent setup is deliberately minimal, consisting of only two agents, where the second agent adapts to the policy of the first by using a sequential learning paradigm. This simplification is sufficient for assessing the research questions of this paper. The modular, decentralized approach is increasingly being explored for industrial control tasks, such as optimizing manufacturing system throughput, as it offers a scalable and robust alternative to training a single, monolithic agent for a complex, multi-faceted problem [6,21].

# ENVIRONMENT DESIGN

The simulation environment presented in this work is designed to model a multi-stage industrial material handling process, providing a benchmark for comparing different reinforcement learning control strategies. For a detailed qualitative analysis of the system's dynamics and the emergent agent behavior, a comprehensive dashboard (see Figure 2) is utilized to observe the environment's state and its dynamics.

Our framework is constructed by combining two complementary benchmarks, SortingEnv and ContainerGym, which were previously integrated into a foundational simulation [7–9]. The key novelty in this work is the extension from a single-agent setup to a multi-agent system. A second learning agent has been introduced to manage the container pressing process, a task that was previously handled by a rule-based heuristic. The environment simulates a sequential material flow, as illustrated in Figure 1, where raw material passes through an input and belt stage, is sorted into one of five containers, and is finally compressed into bales by one of two presses.

This setup is designed to facilitate two distinct experimental paradigms: a decentralized, modular approach, where two separate agents are trained for the sorting and pressing tasks, and a centralized, monolithic approach, where a single agent learns to control both processes concurrently.

# Actions, Observations, and Reward Functions

To accommodate the modular and monolithic training schemes, three distinct agent architectures are defined, each with its own observation and action space (see Table 1). The key components are summarized in Table 1. The observation spaces are continuous and normalized, while the action spaces are discrete. The Sorting agent's observation space includes belt occupancy, material proportions, sorting accuracies, and container purity deviations. Its two actions select the sensor mode, as described in literature [9]. The Pressing agent's observation space comprises normalized container fill levels, fill ratios, materials in the sorting stage, and press timers. Its actions consist of a no-op or activating one of two presses for one of five containers. The Monolithic agent's observation space is a concatenation of the sorting and pressing vectors, while its actions represent a flattened combination of the two sorting modes and all pressing actions.

The reward functions are designed to reflect the distinct operational goals of each sub-task. The structure of these functions is crucial for guiding the agents toward effective policies and is visualized in Figure 3.

The reward for the sorting agent is a dense, state-based signal designed to incentivize the maintenance of high material purity in the containers. It is calculated based on the average deviation of the current purity levels ($p$) from their predefined thresholds ($\theta$). This raw score is then scaled and transformed by a hyperbolic tangent (tanh) function to produce a smooth, bounded reward between -1.0 and 1.0, as shown in Figure 3a. The tanh function creates a sensitive region around the target threshold where small changes in purity result in a significant reward signal, while saturating at the extremes to ensure training stability

**TABLE 1.** Comparison of observation and action space dimensions for the modular and monolithic agents.

| Agent Type | Number of Actions | Number of Observations |
|---|---|---|
| Sorting | 2 | 13 |
| Pressing | 11 | 16 |
| Monolithic | 22 | 29 |

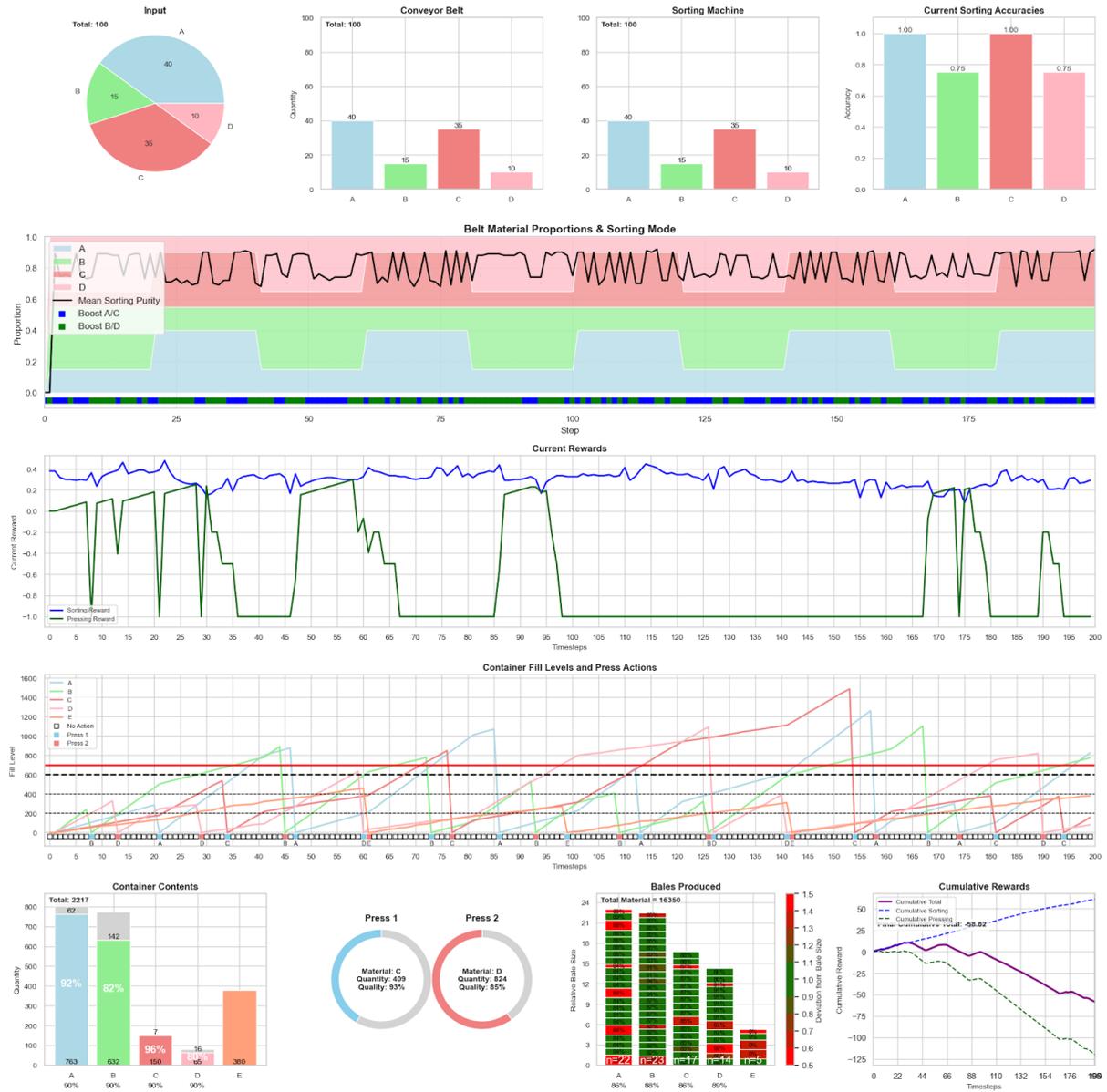

**FIGURE 2.** Comprehensive simulation dashboard illustrating the environment state and agent performance over a 200-timestep episode. The dashboard provides a holistic, multi-faceted view of the system's dynamics. (Top Row) Four plots showing the instantaneous state of the material flow: the stochastic Input composition, material quantities on the conveyor belt and in the sorting machine, and the resulting probabilistic sorting accuracies (time series plots). The subsequent three plots visualize performance over time: belt material proportions are shown alongside the agent's discrete sorting mode decisions; current rewards for both the sorting and pressing agents are tracked individually; and container fill levels for all five containers are plotted against the timeline of discrete press actions (bottom row). Four panels display aggregate metrics at the end of the episode: a breakdown of final container contents including purity percentages; the operational status of the two presses; a summary of all bales produced, indicating their size and quality; and the cumulative reward trajectories for each sub-task and the system as a whole.

The pressing reward is a two-component function designed to balance throughput and efficiency.
1. State-Based Component: A continuous reward is granted at every timestep based on the overall fill ratio of the containers. This component encourages the agent to maintain a high level of material in the system, incentivizing throughput. As shown in Figure 3b, a higher system-wide fill ratio shifts the entire potential reward curve upwards.
2. Action-Based Component: A sparse, event-driven reward is given only when a press action is initiated. This component is structured as a triangular wave, granting the highest reward for pressing integer multiples of the standard bale size (e.g., 1.0, 2.0 bales) and penalizing inefficient actions that result in fractional bales (e.g., 1.5 bales). An additional bonus is incorporated to reward pressing multiple full bales in a single action, which corresponds to efficient operation in practice.

For the monolithic agent, a single, unified reward signal is provided at each timestep. This reward is the unweighted sum of the rewards from the two sub-tasks ($R_{total} = R_{sort} + R_{press}$). This structure challenges the single agent to learn a policy that jointly optimizes both the continuous purity-maintenance task and the sparse, efficiency-driven pressing task.

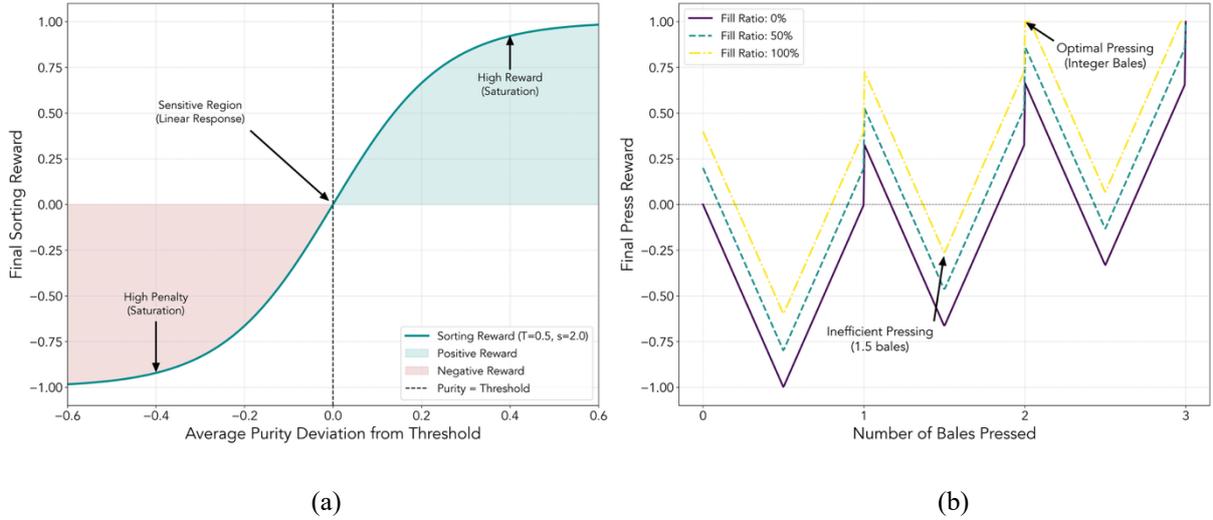

(a) (b)

**FIGURE 3.** Visualization of the reward functions for the modular agents. (a) The state-based Sorting Reward as a function of the average purity deviation from the target threshold. The tanh function creates a sensitive region around the threshold (0.0) and saturates towards +1.0 (high reward) or -1.0 (high penalty) for large deviations. (b) The two-component Pressing Reward as a function of the number of bales produced. The reward's characteristic triangular wave shape, which incentivizes pressing integer numbers of bales, is shifted vertically based on the overall container fill ratio, adding a state-based incentive for maintaining high system throughput.

## EXPERIMENTS AND RESULTS

To evaluate the performance of the modular and monolithic control strategies, we designed a comprehensive training and benchmarking pipeline. All experiments were conducted using the Gymnasium library [22] for the environment backbone and the Stable-Baselines3 [10] framework for the implementation of the reinforcement learning agents.

### Agent Training, Experiments and Benchmarking

The learning algorithm used for all agents in this study is Proximal Policy Optimization (PPO) [23], a state-of-the-art policy gradient method known for its stability and sample efficiency. The PPO agents utilize a policy and value network consisting of a multi-layer perceptron (MLP) with two hidden layers of 32 neurons each. Each agent was trained for a total of 100,000 timesteps, with a maximum of 200 steps per episode, and all training runs were initialized with a fixed random seed (SEED = 42) to ensure reproducibility. In the modular training setup, a sequential, hierarchical approach is employed. First, the Sorting Agent is trained in isolation. The learned policy of this agent is integrated into the Pressing environment providing the sorting decisions. This creates a stable, pre-optimized upstream process, simplifying the Pressing Agent's learning and avoiding the challenge of training two agents simultaneously, where each must adapt to the other's constantly changing policy. Our experimental design involves two distinct conditions to investigate the impact of constraining the action space:

1. Training without Action Masking: In this variant, agents must learn to avoid invalid actions solely through the reward signal. The environment does not provide an explicit negative reward for an invalid action (e.g., attempting to use a busy press); instead, the action is simply ignored, resulting in a wasted timestep. This serves as an implicit penalty by forfeiting the opportunity to earn a potential reward.
2. Training with Action Masking: To quantify the benefits of guiding exploration, a second set of agents was trained using action masking. This technique, implemented via the ActionMasker wrapper and MaskablePPO from the sb3_contrib extension [10], dynamically restricts the agent's policy at each timestep, ensuring that only valid actions are available for selection.

Following the training phase, a rigorous benchmarking process was conducted to compare the performance of the learned policies against two baseline agents: a Random agent that selects actions uniformly (weak baseline) and a Rule-Based agent (strong baseline). The rule-based agent employs a fixed heuristic for each sub-task: the sorting component selects the mode that boosts the most abundant material group currently on the belt, while the pressing component greedily selects the container with the highest fill level as soon as a press becomes available.

The benchmark evaluates five distinct policies: Random, Rule-Based, the trained PPO Sorting agent (paired with the rule-based presser), the fully modular PPO agents (Sort + Press), and the monolithic PPO agent. To ensure statistical significance, the performance of each policy was evaluated across 10 different, unseen environment seeds. Each evaluation episode was run for 200 timesteps, with the mean and standard deviation of the cumulative reward serving as the primary comparison metric.

## Results

The performance of the modular and monolithic agent configurations was evaluated under two distinct experimental conditions: with and without the use of action masking. The benchmark results, presented as the mean cumulative reward over 10 independent seeds, reveal the significant impact of action space constraints on policy learning.

The benchmark results for agents trained without the action masking mechanism are presented in Figure 4. A remarkable outcome is the strong performance of the Rule-Based policy, which achieved a high positive cumulative reward, establishing a robust baseline. In stark contrast, all trained RL agents failed to learn a beneficial policy, scoring significantly negative rewards and performing substantially worse than the rule-based heuristic. All learning-based agents did, however, outperform the Random agent baseline.

Among the learning-based approaches, a clear trend emerged favoring modularity. The "Sort Agent," when paired with a rule-based presser, outperformed the fully monolithic "Combined Agent." The strongest performance among the learning agents was achieved by the fully modular "Sort + Press Agents" configuration, which demonstrated that combining two individually trained agents was superior to using a single trained agent in this unconstrained setting. When action masking was enabled during training, a dramatic improvement in performance was observed across all RL agents, with every learning-based strategy outperforming the random heuristic. In this masked environment, the performance gap between the modular and monolithic approaches narrowed considerably. The fully modular "Sort + Press Agents" and the monolithic "Combined Agent" achieved a similar level of performance, with the monolithic agent showing a slight advantage. Despite these marked improvements, the Rule-Based system remained the strongest overall policy, still achieving a higher cumulative reward than any of the trained RL agents.

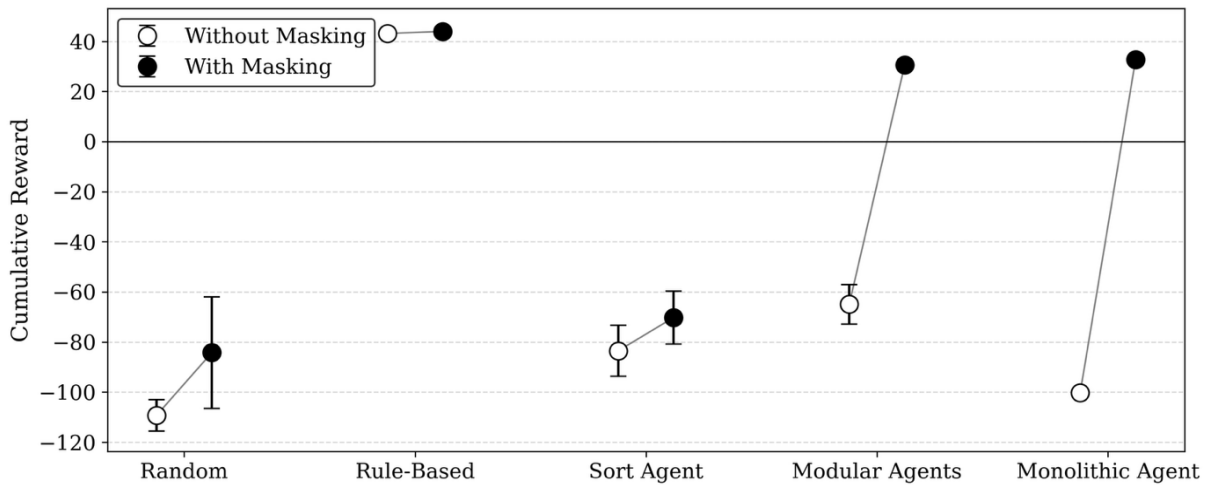

**FIGURE 4.** Agent Performance Comparison with and without Action Masking. The plot shows the mean cumulative reward across 10 independent seeds for each agent type. Error bars indicate standard deviation. Across all learning-based agents, action masking yields a clear performance improvement, with modular and monolithic agents achieving positive rewards.

# DISCUSSION

This study addressed the challenge of applying reinforcement learning to complex, multi-stage industrial processes. We introduced a novel benchmark environment that simulates a sequential waste sorting and pressing workflow by combining two existing benchmarks [7–9]. The experiments conducted within this new environment provide several key insights into agent architecture, training strategies, and the standing of current RL methods against traditional heuristics.

The experiments confirm that action masking is an effective mechanism for improving training efficiency and stabilizing policy learning in domains with invalid or redundant actions. Similar findings have been reported in other scenarios, where masking reduces exploration overhead and accelerates convergence [24]. In our benchmark, the addition of masking led to an improvement in cumulative rewards across all agents.

We also contribute to the debate on specialization versus integration in multi-agent reinforcement learning. Without masking, modular agents outperformed the monolithic agent, consistent with earlier results in coordination-heavy benchmarks such as SMAC, where decentralized specialization mitigates the burden of learning in large joint action spaces [25]. When masking was applied, however, the performance gap between modular and monolithic strategies largely disappeared, suggesting that constraints on the action space reduce the difficulty of the learning problem to the point where a centralized policy can be competitive.

A further remarkable result is the strength of the rule-based baseline, which consistently outperformed all learning-based strategies. This highlights the ongoing gap between reinforcement learning methods and carefully designed heuristics in highly structured industrial environments. The outcome aligns with current industrial practice, where heuristics remain dominant due to their interpretability, reliability, and ease of deployment [4]. RL approaches, in contrast, are still in early stages of benchmarking and rarely reach production-grade performance in real plants [4,12].

Despite these contributions, the present study has several limitations. The simulation environment omits physical stochasticity and sensor noise, both of which are central to real-world industrial processes [17]. The reward functions, while interpretable, are simplified and rely on strong assumptions about task objectives. Training was performed for a relatively small number of timesteps (100,000), which may not allow policies to fully exploit the structure of the environment. Moreover, generalization beyond the tested scenario was not investigated. Finally, the rule-based baseline may be particularly well suited to the design of this environment, possibly overstating its advantage over RL.

Future work should extend the benchmark with more realistic process models, including stochasticity and disturbances, and explore more advanced RL techniques, such as curriculum learning and hybrid approaches that integrate expert knowledge, to develop more robust and practical solutions for industrial automation.

# CONCLUSION

This study introduced an industry-inspired benchmark to investigate the trade-offs between modular and monolithic RL agent architectures in a multi-stage industrial process. Our central finding is that the choice between a specialized, modular design and a centralized, monolithic one is heavily dependent on the complexity of the action space. While specialized agents learn more effectively in unconstrained environments, a monolithic agent can achieve comparable performance once the action space is simplified using techniques such as action masking. This suggests that the key to success for centralized agents in this case lies in effective action space management, rather than inherent difficulties in coordinating multiple tasks. Across all settings, a simple rule-based heuristic remained the strongest competitor, highlighting the significant challenge for RL to surpass well-engineered traditional solutions in structured domains. This work provides a valuable testbed for future research aimed at closing this gap.

## ACKNOWLEDGMENTS
Funded by the German Federal Ministry for Economic Affairs and Climate Action through the grant "EnSort".

## CODE AVAILABILITY
The code is available here.